\documentclass[runningheads]{llncs}
\usepackage{graphicx}
\usepackage{caption}
\usepackage{subcaption}
\usepackage{amsmath}
\usepackage{comment}
\usepackage{mathrsfs}
\usepackage{algorithmic}[1]
\usepackage{algorithm}
\usepackage{mathtools}

\DeclarePairedDelimiter\floor{\lfloor}{\rfloor}
\begin{document}
\title{Multi-Robot Path Planning Via Genetic Programming}
%
%
\author{Alexandre Trudeau \and
Christopher M. Clark}
\authorrunning{Trudeau \& Clark}
%
\institute{Harvey Mudd College, Claremont CA 91711, USA \\
\email{\{atrudeau,clark\}@hmc.edu}}
\maketitle              
\begin{abstract}

This paper presents a Genetic Programming (GP) approach to solving multi-robot path planning (MRPP) problems in single-lane workspaces, specifically those easily mapped to graph representations. GP's versatility enables this approach to produce programs optimizing for multiple attributes rather than a single attribute such as path length or completeness. When optimizing for the number of time steps needed to solve individual MRPP problems, the GP constructed programs outperformed complete MRPP algorithms, i.e. Push-Swap-Wait (PSW), by $54.1\%$. The GP constructed programs also consistently outperformed PSW in solving problems that did not meet PSW's completeness conditions. Furthermore, the GP constructed programs exhibited a greater capacity for scaling than PSW as the number of robots navigating within an MRPP environment increased. This research illustrates the benefits of using Genetic Programming for solving individual MRPP problems, including instances in which the number of robots exceeds the number of leaves in the tree-modeled workspace.

\end{abstract}

\section{Introduction}
Multi-robot systems offer greater performance than single robot systems at the cost of addressing issues, including the potential for collisions, bottlenecks, and traffic jams that can occur when many robots are navigating within confined or single-lane environments. To address such issues, the problem of Multi-Robot Path Planning (MRPP) has been explored within many contexts, e.g. \cite{Cao1997,clark2004,Dudek1996}. A solution to the MRPP problem typically requires the construction of a collision-free path for each robot having a unique starting location and unique goal location within the workspace. In solving this problem, researchers have focused on developing algorithms that are scalable, complete, and/or optimal (e.g. with respect to path length) \cite{Carpin02onparallel,Luna2011,Luna:2011:PSF:2283396.2283446,clark2008,korf2011,Wang2014}.

\begin{figure}
\centering
\begin{subfigure}[b]{0.5\textwidth}
\includegraphics[height=1.2in]{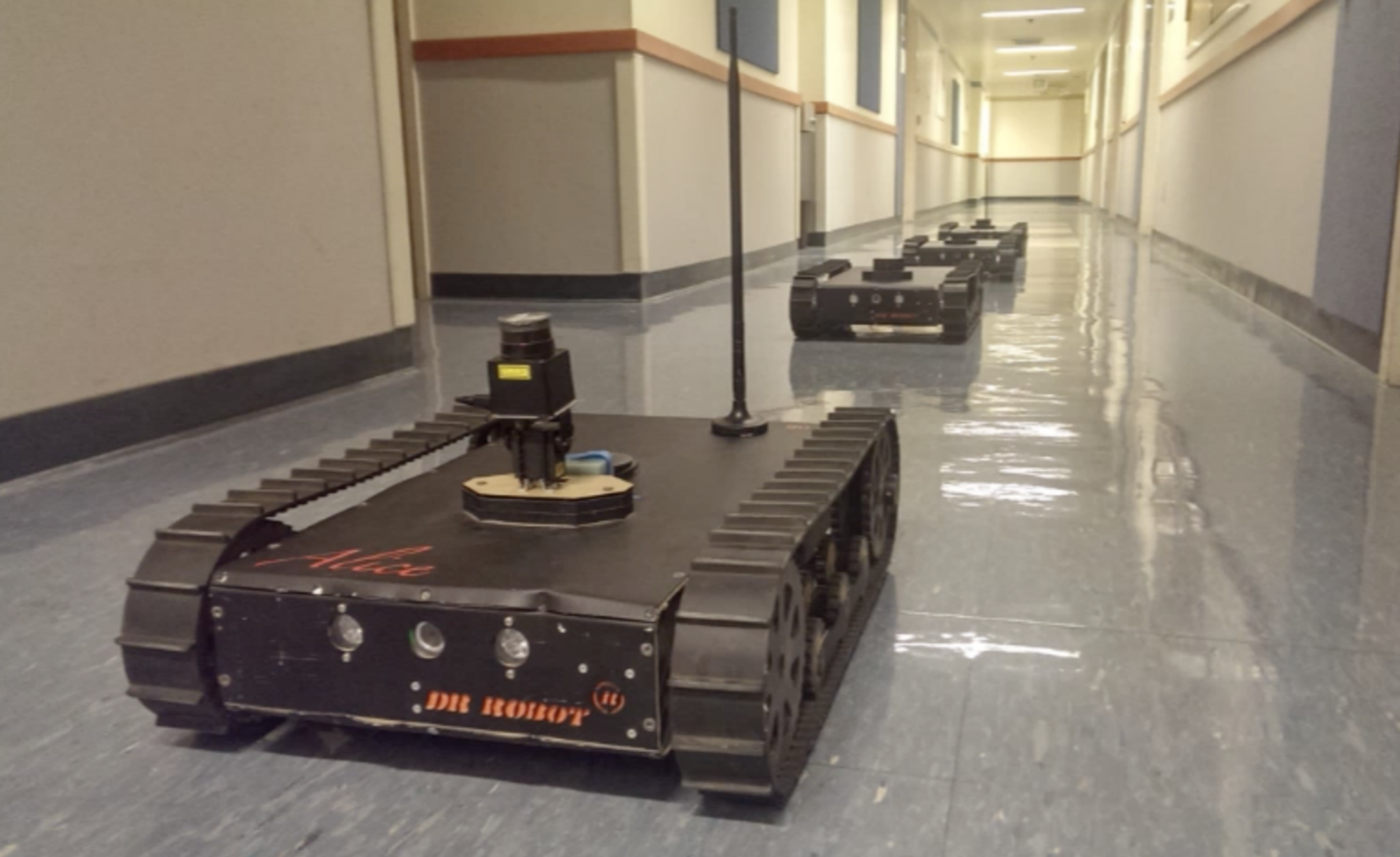}
\caption{}
\label{fig1a}
\end{subfigure}
\begin{subfigure}[b]{0.5\textwidth}
\includegraphics[height=1.2in]{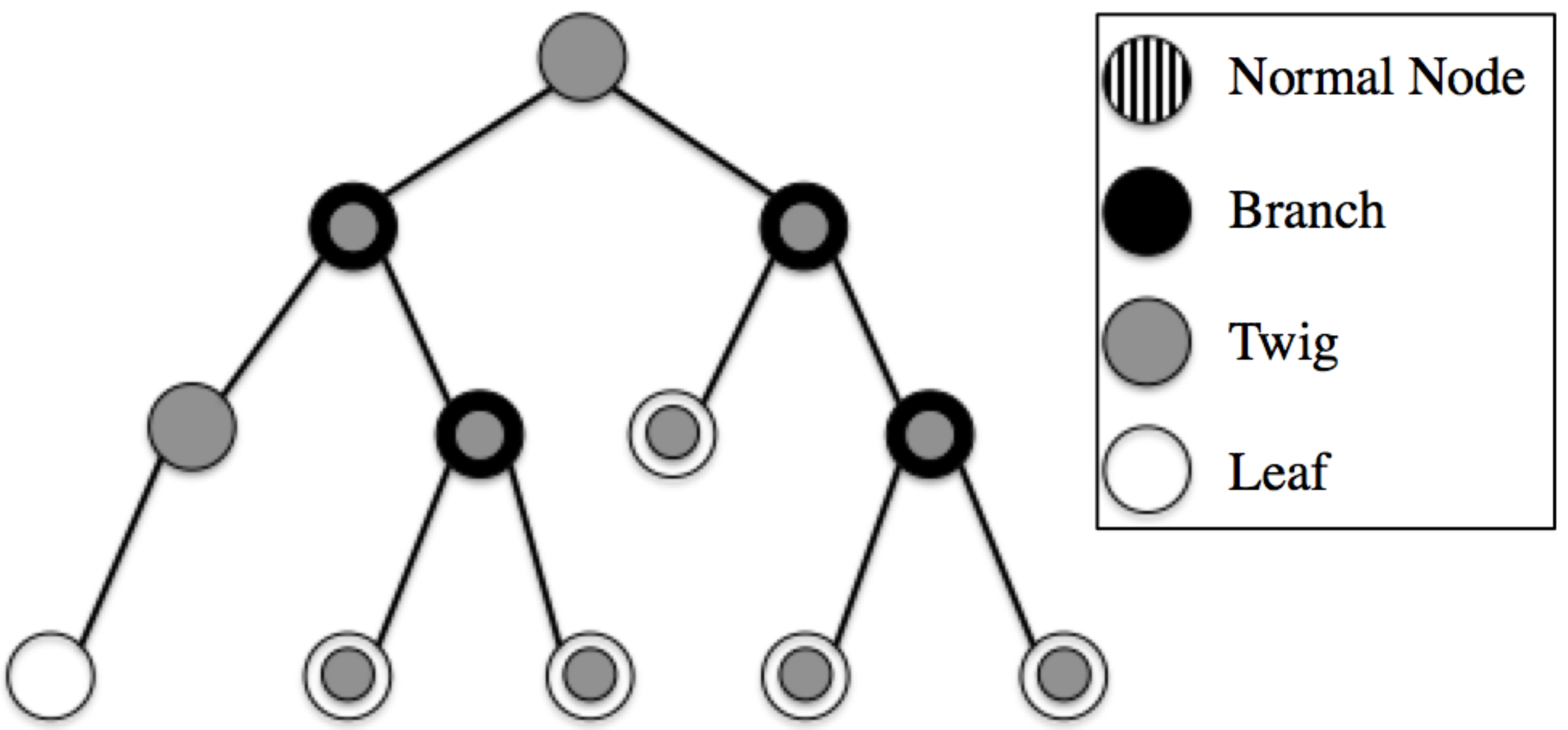}
\caption{}
\label{fig1b}
\end{subfigure}
\hspace{10pt}
\begin{subfigure}[b]{0.5\textwidth}
\includegraphics[height=1.2in]{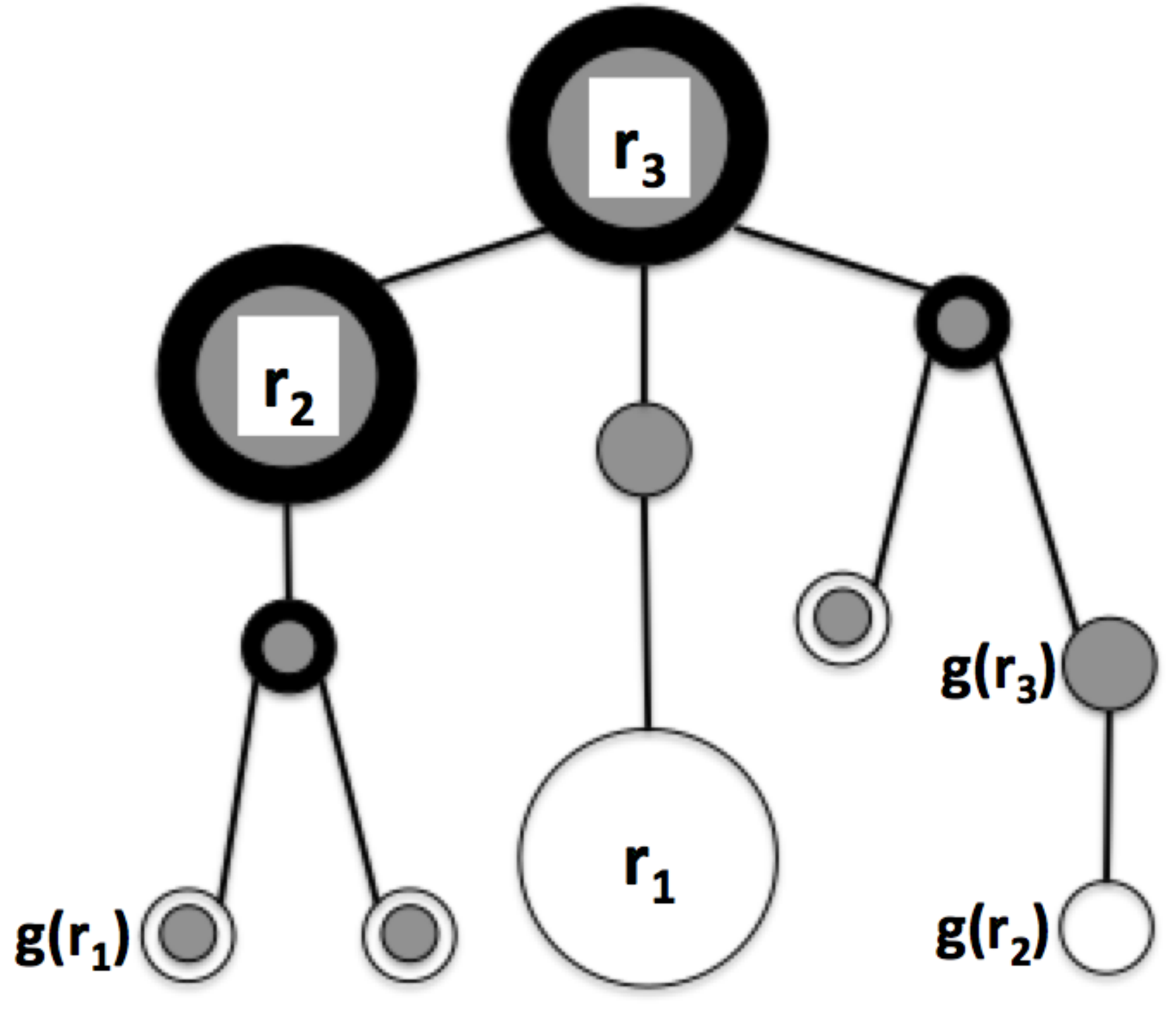}
\caption{}
\label{fig1c}
\end{subfigure}
\caption{Example of a single-lane tunnel/corridor environment (a). Example of a single-lane tunnel/corridor environment being represented as a minimum spanning tree (b). In (c), a randomly generated MRPP scenario. } 
\end{figure}

This paper focuses on MRPP within the single-lane environments typically found in mines and warehouses. In these workspaces, fleets of autonomous robots navigate single lane tunnels; for example, in Fig. \ref{fig1a}, the robots do not have space to move around each other safely. As in previous work, this research abstracts such workspaces as minimum spanning trees (see Fig. \ref{fig1b}), in which robot actions correspond to movement along edges connecting nodes \cite{references:PSW_IROS}.

This paper presents a Genetic Programming (GP) approach to single-lane MRPP. GP is an evolutionary algorithm that evolves computer programs as solutions to problems. Here, a centralized master computer first runs the GP algorithm offline on a training set of MRPP examples, learns a computer program that solves these examples, and shares this program with the robots within the environment. Then, the robots sequentially execute the learned program in real-time to determine their next actions. Each robot runs this computer program at each time step until all of the robots have visited their respective goal destinations. The specific contributions of this paper include:
\begin{enumerate}
    \item Identification of a set of GP functions and terminals for generating MRPP programs.
    \item A GP algorithm for constructing programs solving MRPP problems.
    \item Simulation results illustrating GP performance in solving single MRPP instances and in developing a general MRPP program.
\end{enumerate}


In Section 2, MRPP/GP related research is introduced. Section 3 provides the problem formulation. In Section 4, a GP algorithm is outlined and its application to MRPP explained. In Section 5, the computational complexity of the GP algorithm is evaluated. Section 6 describes how the GP's function set and terminal set were developed. Section 7 documents the experiments performed to validate the approach. In Section 8, conclusions are presented and future work is considered.

\section{Background}

MRPP algorithms can be classified in multiple ways. First, they can be classified as \textit{centralized} or \textit{decentralized}. Centralized MRPP algorithms have a single robot or central computer plan the paths of all the robots \cite{Carpin02onparallel,Luna2011,Luna:2011:PSF:2283396.2283446,clark2008}. Decentralized MRPP algorithms have each individual robot plan its own path \cite{kumar2008,clark2004,luna_thesis,velagapudi2010,zheng2008}. MRPP algorithms can also be distinguished by whether they are \textit{local} or \textit{global}. Local MRPP algorithms determine a robot's next move at each time step whereas global MRPP algorithms generate a robot's entire path before it sets off towards its destination. 

Related to multi-robot path planning in graphs is the work on pebble motion. A pebble motion algorithm aiming to move pebbles around a tree from a start to goal configuration is shown in \cite{auletta}. Earlier work also includes \cite{references:kornhauser}. More recent work \cite{references:yu} presents an algorithm that upper bounds the number of moves required for moving pebbles around a graph to their goal configuration. Finally, a polynomial time algorithm for  coordinating the motion of labeled discs in high disc density scenarios \cite{chinta} is shown to have optimality guarantees. 

Recent centralized MRPP approaches include \cite{references:han}, a polynomial time algorithm (SEAR) with expected optimality guarantees in obstacle-free environments. In \cite{references:yuB}, MRPP algorithms for optimizing across several metrics (e.g. makespan, max distance, etc.) are shown to calculate near optimal paths for 100+ robots in seconds. Other work \cite{references:ma} aims to maximize the number of agents that can reach their goal under a deadline.

Presented in \cite{references:PSW_IROS}, the Push-Swap-Wait algorithm (PSW) is a decentralized, local, and topology-based single-lane MRPP algorithm that distinguishes itself from other MRPP algorithms in that it is \textit{complete}: it is guaranteed to solve MRPP problems that meet certain conditions. Unlike most multi-agent path finding approaches, however, PSW considers an environment solved once each robot has visited its unique destination. Under this definition, robots do not need to remain at their destination once solved. Centralized and complete MRPP algorithms have also been developed \cite{clark2008}.

GP has been successfully used to evolve collision avoidance programs for single robots. The collision avoidance programs generated by the GP in \cite{references:Nordin97realtime} are structured as trees containing arithmetic and boolean operators. The programs take in robot sensor readings and return motor speeds.

GP has also been used in MRPP. Kala used Grammar Guided Genetic Programming (GGGP) to evolve optimal paths for individual MRPP examples \cite{references:Kala_GGGP}. For a given instance of MRPP, GGGP is run for each robot so that they learn the best paths to their respective destinations. A master genetic algorithm then optimizes across robots, selecting the path for each robot ensuring that the robots collectively reach their destinations as quickly as possible without colliding. 

Evolutionary algorithms other than GP have been used in MRPP. In \cite{references:Wang_Wu} and \cite{references:Coevolution_CGA}, Cooperative Co-evolution is used to generate paths that achieve collision avoidance in individual MRPP examples. Here, the entities undergoing evolution are the paths themselves, which is useful in producing environment specific optimal paths. Chakraborty, Konar, Jain, \& Chakraborty use Differential Evolution to generate algorithms for MRPP \cite{references:Differential_Evolution}. In this approach, each robot has its own path-planning algorithm that evolves. Lastly, Das, Sahoo, Behera, \& Vashisht use Particle Swarm Optimization to generate decentralized and local algorithms for MRPP  \cite{references:MRMP_PSO}.

This paper presents a decentralized, local, and topology-based GP approach to single-lane MRPP. The performance of this approach is compared to PSW. Similarly to PSW, an environment is considered solved once each robot has visited its destination. Unlike PSW and the evolutionary algorithms mentioned above, the GP approach presented in this paper is the first research we know of that constructs decentralized and scalable single-lane MRPP programs using GP.

\section{Problem Formulation}
\begin{figure}
\includegraphics[width=\textwidth]{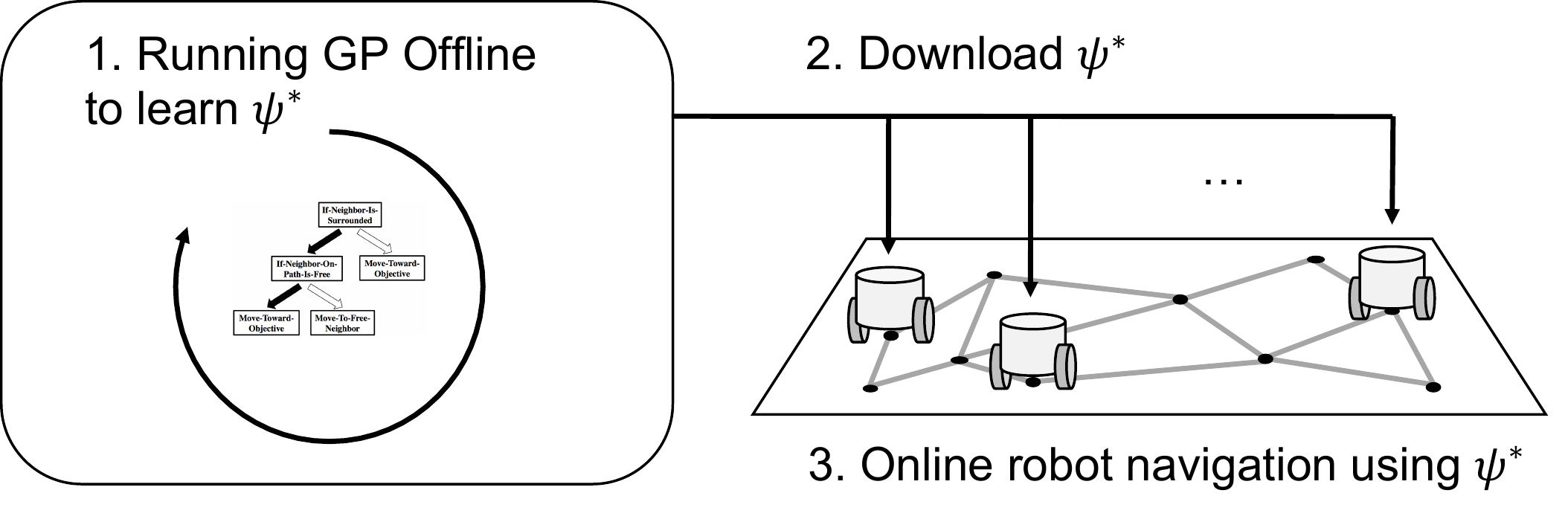}
\caption{Flowchart showing the interaction between offline and online computation.} \label{figGPflow}
\end{figure}

This research attempts to develop a GP algorithm that can be run offline to construct a solution program $\psi *$ to be downloaded to multiple individual robots (see Fig. \ref{figGPflow}). Then, each robot can execute $\psi *$ to enable collision-free, decentralized navigation through a given single-lane workspace to its goal destination.

In designing the GP algorithm that constructs $\psi *$, it is assumed that the navigable portions of the workspace are modeled as a fully connected graph, which can in turn be represented with a minimum spanning tree $T(N,E)$, composed of a set of nodes, $N$, connected with a set of undirected edges, $E$. Recall that a minimum spanning tree $T(N,E)$ of graph $G(N,\varepsilon)$ contains the same set of nodes $N$ and contains a subset of the edges $E \subseteq \varepsilon$ such that all of the nodes remain connected and there are no cycles. The tree $T$ is occupied by a set of robots $R$ that are assumed to travel at the same constant speed. Each robot may traverse an edge during a time step and must occupy a node at the end of a time step. Only a single robot may occupy a given node at a time. 

At each time step $t$, each robot $r_i \in R$ runs the GP generated program $\psi *$, which returns the robot's assignment $A(r_i,t) \in N$. Thus, if $A(r_i,t-1) = n_1$ and $A(r_i,t) = n_2$, robot $r_i$ traverses edge $e_{n_1 \rightarrow n_2}$ during time step $t$ if $n_1 \neq n_2$ and $e_{n_1 \rightarrow n_2} \in E$. Each edge $e \in E$ is of the same length and takes a single time step to traverse. An edge can only be traversed by a single robot during a given time step. Each robot $r$ starts on a unique starting node $s(r) \in N$ such that $s(r_i) \neq s(r_j)$ if $r_i \neq r_j$. Each robot also has a unique destination node $g(r) \in N$ such that $g(r_i) \neq g(r_j)$ if $r_i \neq r_j$.

All robots within $\rho$ nodes of $r_i$ in the minimum spanning tree $T(N,E)$ are considered to be in $r_i$'s direct communication network: $c(r_i)$.

Finally, a MRPP problem is considered solved at time $t_s$ if $\forall r_i \in R, \exists t_i | t_i \leq t_s$ and $A(r_i,t_i) = g(r_i)$. In other words, a MRPP problem is considered solved once each robot has visited its destination node.

\section{Genetic Programming in MRPP}

Genetic Programming is an evolutionary algorithm that begins with an initial, randomly generated population of solution programs. These programs are iteratively tested against a set of training examples to evaluate their fitness---how well they solve the training problems.

\begin{table}
\caption{The GP function set for MRPP.}\label{tab1}
\begin{tabular}{
  @{\kern-.5\arrayrulewidth}
  |p{\dimexpr3.0cm-2\tabcolsep-.5\arrayrulewidth}
  |p{\dimexpr8.3cm-2\tabcolsep-.5\arrayrulewidth}
  |@{\kern-.5\arrayrulewidth}
}
\hline
{\bfseries Function} & {\bfseries Description} \\
\hline
\textbf{If-Two-Robots-On-Each-Others-Path} &
Checks whether the current robot $r_i$'s position $A(r_i, t)$ is on the path of another robot $r_j$ that belongs to $c(r_i)$. Also checks whether $A(r_j, t)$ is on $r_i$'s path. If both conditions are met, $r_i$ and $r_j$ each set their paths to the nearest branch node to $r_i$.\\
\hline
\textbf{If-Neighbor-Is-Surrounded} &
Checks whether the current robot $r_i$ is adjacent to another robot $r_j$ that is surrounded---that is, each of $r_j$'s adjacent nodes are occupied by a robot.\\ 
\hline
\textbf{If-Robot-At-Branch} &
Checks whether the current robot $r_i$ is at a branch node that it set its path to.\\
\hline
\textbf{If-Robot-At-Destination} &
Checks whether the current robot $r_i$ is at its destination node $g(r_i)$.\\
\hline
\textbf{If-Robot-Moving-To-Branch} &
Checks whether the current robot $r_i$ is travelling to a branch node it has set its path to but hasn't reached yet.\\
\hline
\textbf{If-Neighbor-On-Path-Is-Free} &
Checks whether the next node on the current robot $r_i$'s path is free.\\
\hline
\textbf{If-Robot-Is-Solved} &
Checks whether the current robot $r_i$ has visited its destination node $g(r_i)$.\\
\hline
\textbf{If-On-Path-Of-Robot-In-Network} &
Checks whether the current robot $r_i$'s position $A(r_i, t)$ is on the path of another robot $r_j$ that belongs to its direct communication network $c(r_i)$.\\
\hline
\textbf{If-Robot-In-Network-Moving-To-Branch} &
Checks whether a robot $r_j$ that belongs to $c(r_i)$ is travelling to a branch node.\\
\hline
\end{tabular}
\end{table}

The next generation of programs is then produced via asexual reproduction, genetic crossover, and mutation, whereby more fit solutions are granted a greater likelihood of surviving the evolutionary process. This leads the population of solutions to become increasingly fit over the course of generations \cite{Koza:1992:GPP:138936}.

\subsection{Solution Representation}
GP solutions are computer programs that are structured as trees. The internal nodes in the tree-structured programs are sampled from the function set $F = \{f_1,f_2,...,f_N\}$, where a function can be an arithmetic/boolean/conditional operator, mathematical/iterative/recursive function, etc. Here, $F$ includes conditional if-statements relying on information about the environment: e.g. the positions of robots relative to other robots, the positions of robots relative to branch nodes, etc. (see Table 1). Whenever a function or terminal requires the current robot to compute a path from its current node to its destination node or to a branch node, Dijkstra's algorithm is used. To note, a branch node is a node in the minimum spanning tree $T(N,E)$ having 3 or more neighbors.

The leaves in the computer program's tree structure are sampled from the terminal set $L = \{l_1,l_2,...,l_N\}$ and can be constants, variables, or functions changing the state of the environment. The terminals used here are actions dictating a robot's movement between nodes in the workspace (see Table 2). Each of the terminals contain a block of code that checks whether execution of the terminal would result in a collision. If a robot executes a terminal that would cause it to travel to a node that is already occupied by another robot, this block of code ensures that the robot stays at its current node in order to prevent a collision.

Given the programs' internal nodes are conditional operators, the resulting solution programs represent decision trees solving MRPP problems (e.g. Fig. 3a-b).

\begin{table}
\caption{The GP terminal set for MRPP.}\label{tab1}
\begin{tabular}{
  @{\kern-.5\arrayrulewidth}
  |p{\dimexpr3.0cm-2\tabcolsep-.5\arrayrulewidth}
  |p{\dimexpr8.3cm-2\tabcolsep-.5\arrayrulewidth}
  |@{\kern-.5\arrayrulewidth}
}
\hline
{\bfseries Terminal} & {\bfseries Description} \\
\hline
\textbf{Move-Toward-Branch} &
The current robot $r_i$ moves to the next node on the path to the branch it has set its path to.\\
\hline
\textbf{Move-To-Free-Neighbor} &
The current robot $r_i$ moves to an adjacent node that isn't occupied by a robot and that it hasn't visited before.\\
\hline
\textbf{Move-Toward-Objective} &
The current robot $r_i$ moves to the next node on its current path.\\
\hline
\textbf{Stay} &
The current robot $r_i$ does not move during the time step and stays at the same node.\\
\hline
\end{tabular}
\end{table}

\subsection{GP Algorithm}

\begin{algorithm}
\hspace*{\algorithmicindent} \textbf{Inputs}: $|P|$, $i_r$, $i_g$, $X$, $p_a$, $p_c$, $p_m$
\begin{algorithmic}[1] 
  \caption{Genetic Programming Adapted for MRPP}
    \STATE $\tau_{B} \leftarrow \infty$\\
    \STATE $\psi* \leftarrow \texttt{None}$\\
    \FOR{$i \leftarrow 1$ \TO $i_r$}
        \STATE $P \leftarrow getInitialPopulation(|P|)$\\
        \FOR{$j \leftarrow 1$ \TO $i_g$}
            \FOR{$k \leftarrow 1$ \TO $|P|$}
                \STATE $\mathscr{F}_k \leftarrow calculateFitness(\psi_k, X)$
                \IF{$\mathscr{F}_k = 0$ \bf{and} $\sum_{n=1}^{|X|} \tau_{k,n} < \tau_{B}$}
                    \STATE $\tau_{B} \leftarrow \sum_{n=1}^{|X|} \tau_{k,n}$
                    \STATE $\psi* \leftarrow \psi_k$
                \ENDIF
            \ENDFOR
            \STATE $P_a \leftarrow asexualReproduction(P, p_a)$ \\
            \STATE $P_c \leftarrow geneticCrossover(P, p_c)$ \\
            \STATE $P_m \leftarrow mutation(P, p_m)$ \\
            \STATE $P \leftarrow P_a \cup P_c \cup P_m$
        \ENDFOR
    \ENDFOR
    \RETURN $\psi*$
\end{algorithmic}
\end{algorithm}

\noindent
In Alg. 1, the GP evolves solution programs solving MRPP problems over the course of $i_r$ runs, each lasting $i_g$ generations. At the start of each run, $getInitialPopulation(|P|)$ is executed (line 4), randomly generating an initial population of $|P|$ programs, with maximum tree depth 2, composed of the functions and terminals from Table 1 and Table 2. Then, for each generation within the given run, the GP iterates through each program $\psi_k$ in the population $P$ and executes $calculateFitness(\psi_k, X)$ (line 7). This function runs a given program $\psi_k$ against a set of MRPP training problems $X$. The time $\tau_{k,n}$ it takes program $\psi_k$ to solve training problem $x_n$ is recorded for all programs in $P$ over all examples in $X$. If a program $\psi_k$ solves all of the examples in $X$ in fewer time steps than the current best program $\psi*$, then the minimum number of time steps taken to solve the examples in $X$, $\tau_{B}$, is updated (line 9) and $\psi_k$ becomes the new best program $\psi*$ (line 10). At the end of each generation, the next generation's population of programs $P$ is produced by running $asexualReproduction(P, p_a)$ with asexual reproduction rate $p_a$ (line 13), $geneticCrossover(P, p_c)$ with genetic crossover rate $p_c$ (line 14), and $mutation(P, p_m)$ with mutation rate $p_m$ (line 15). $p_a$, $p_c$, and $p_m$ define the proportion of programs in the next generation that come from asexual reproduction, genetic crossover, and mutation, respectively. Thus, $p_a$, $p_c$, and $p_m$ must sum to 1 so that $|P|$ is constant over the generations. This iterative process continues for $i_r$ runs of $i_g$ generations. Once the final run has ended, the best solution program $\psi*$ is returned (line 19).

\subsection{Calculating Fitness}
To evaluate the quality of a program $\psi_k$, it is tested against a set of $|X|$ examples called the \textit{fitness set}: $X = \{x_1, x_2,...,x_{|X|}\}$. Each example $x_i$ is a randomly generated MRPP problem including a minimum spanning tree $T$ and a set of robots $R$ having unique starting and destination nodes. 

Notably, a program $\psi_k$ is tested against each example $x_i$ until the example has been solved or the maximum allowable time steps $M$ has been exceeded, where $M = |N|^2*|R|^2$, $|N|$ is the number of nodes in $T$, and $|R|$ is the number of robots. The fitness function to be $minimized$ is defined as $\mathscr{F}_k= \sum_{i=1}^{|X|} f_{k,i}$. 

Specifically, program $\psi_k$'s fitness for training problem $x_i$ is:

\begin{equation}
f_{k,i}=
   \begin{cases}
    0 & x_i \text{ is solved}\\
    \sum\limits_{j=1}^{|R|} dist(A(r_j,M),g(r_j))^2 & \text{otherwise}
\end{cases} 
\end{equation}

\noindent
where $dist(A(r_j,M),g(r_j))$ returns the number of edges along the shortest path from $r_j$ to $g(r_j)$ at time $M$. A training problem $x_i$ can be considered solved but not have $\sum\limits_{j=1}^{|R|} dist(A(r_j,M),g(r_j))^2 = 0$ because a problem is considered solved once each robot has visited its destination. Robots do not need to remain at their destinations once they have been visited.

To evolve MRPP programs navigating robots to their destination nodes in as few time steps as possible, the fitness function calculation only allows a candidate program a total of $\tau_{B}$ time steps to solve all of the examples in $X$. As the GP evolves, $\tau_{B}$ is updated to be the minimum number of time steps that has been found to solve $X$ (line 9).

\subsection{Generating New Populations}
Once the fitness score has been calculated for each program in the population, the next generation's population of programs is created---$p_a\%$ from asexual reproduction, $p_c\%$ from genetic crossover, and $p_m\%$ from mutations. Programs are selected to undergo these operations with fitness proportionate probability $\alpha_k = \beta_k / \sum_{j=1}^{|P|}\beta_j$, where $\beta_k = 1/(1+\mathscr{F}_k)$.

\subsubsection{Asexual Reproduction}
Each program $\psi_k$ is copied from the current population into the new population with probability $\alpha_k$.

\subsubsection{Genetic Crossover}
Two programs from the current population are chosen with fitness proportionate probabilities. Then, a crossover point (i.e. a node) is randomly chosen in both program's decision trees. The subtrees at the crossover points are then swapped and reattached at the crossover points to form two new programs (e.g. Fig. 3c-d) that are added to the new population.

\subsubsection{Mutation}
A program $\psi_k$ from the current population is selected with probability $\alpha_k$. Then, as in genetic crossover, a node in the program's decision tree is uniformly randomly chosen. The chosen node and its subtree are deleted and replaced with a randomly generated subtree (e.g. Fig. 3d). 

\begin{figure}
\centering
\begin{subfigure}[l]{0.45\textwidth}
\includegraphics[height=1.1in]{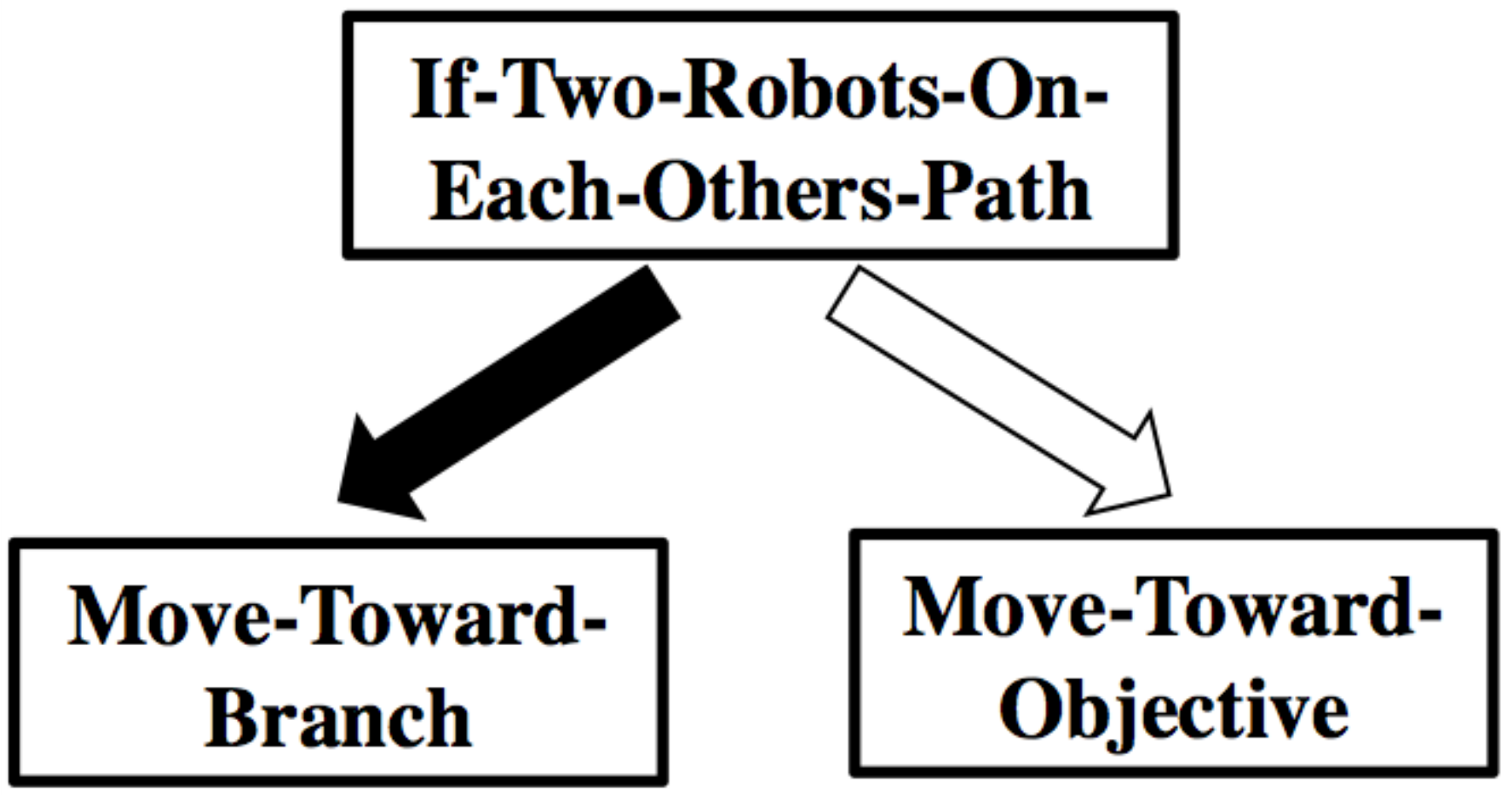}
\caption{}
\end{subfigure}
\begin{subfigure}[l]{0.5\textwidth}
\includegraphics[height=1.1in]{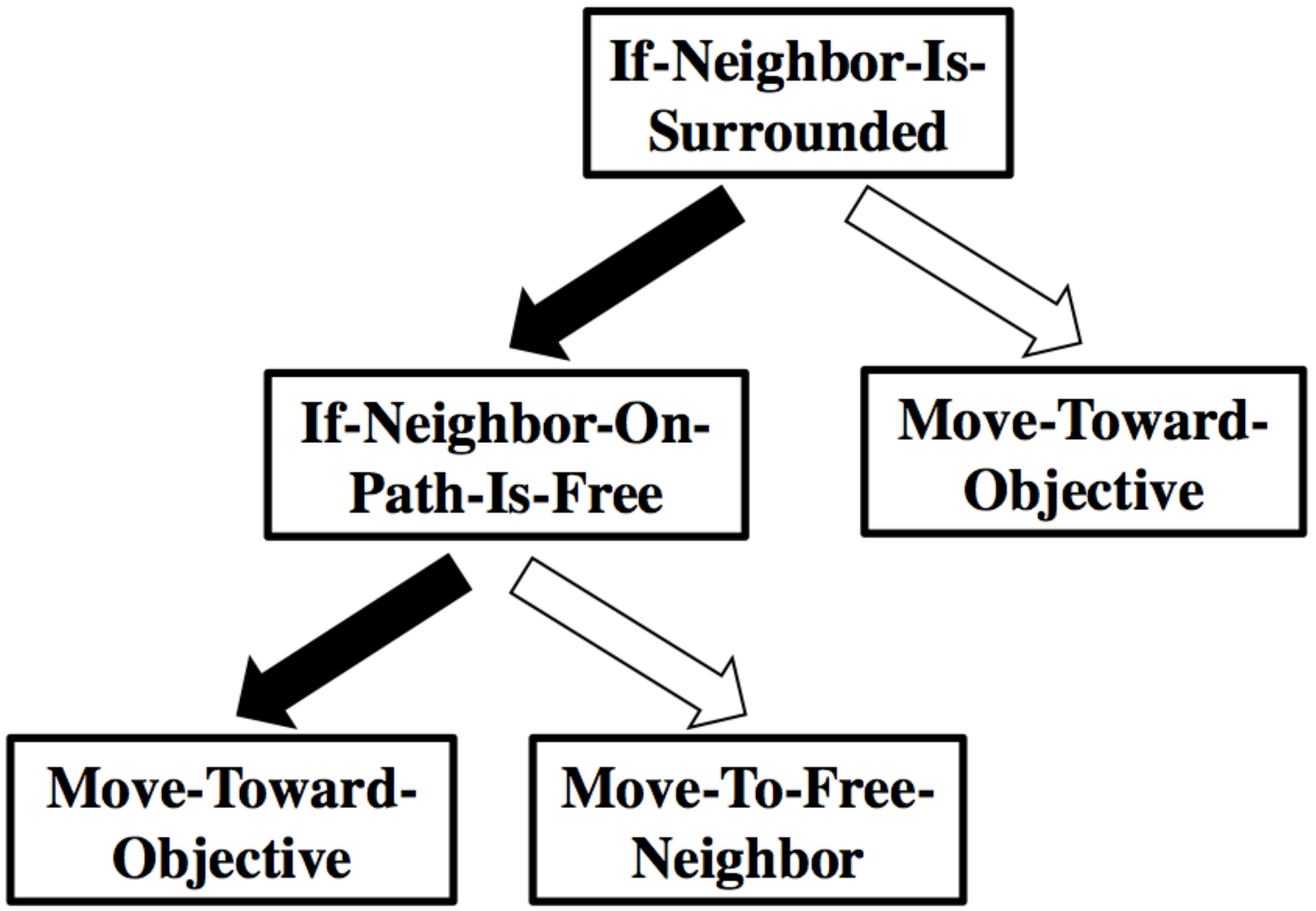}
\caption{}
\end{subfigure}
\hspace{10pt}
\begin{subfigure}[l]{0.45\textwidth}
\includegraphics[height=1.1in]{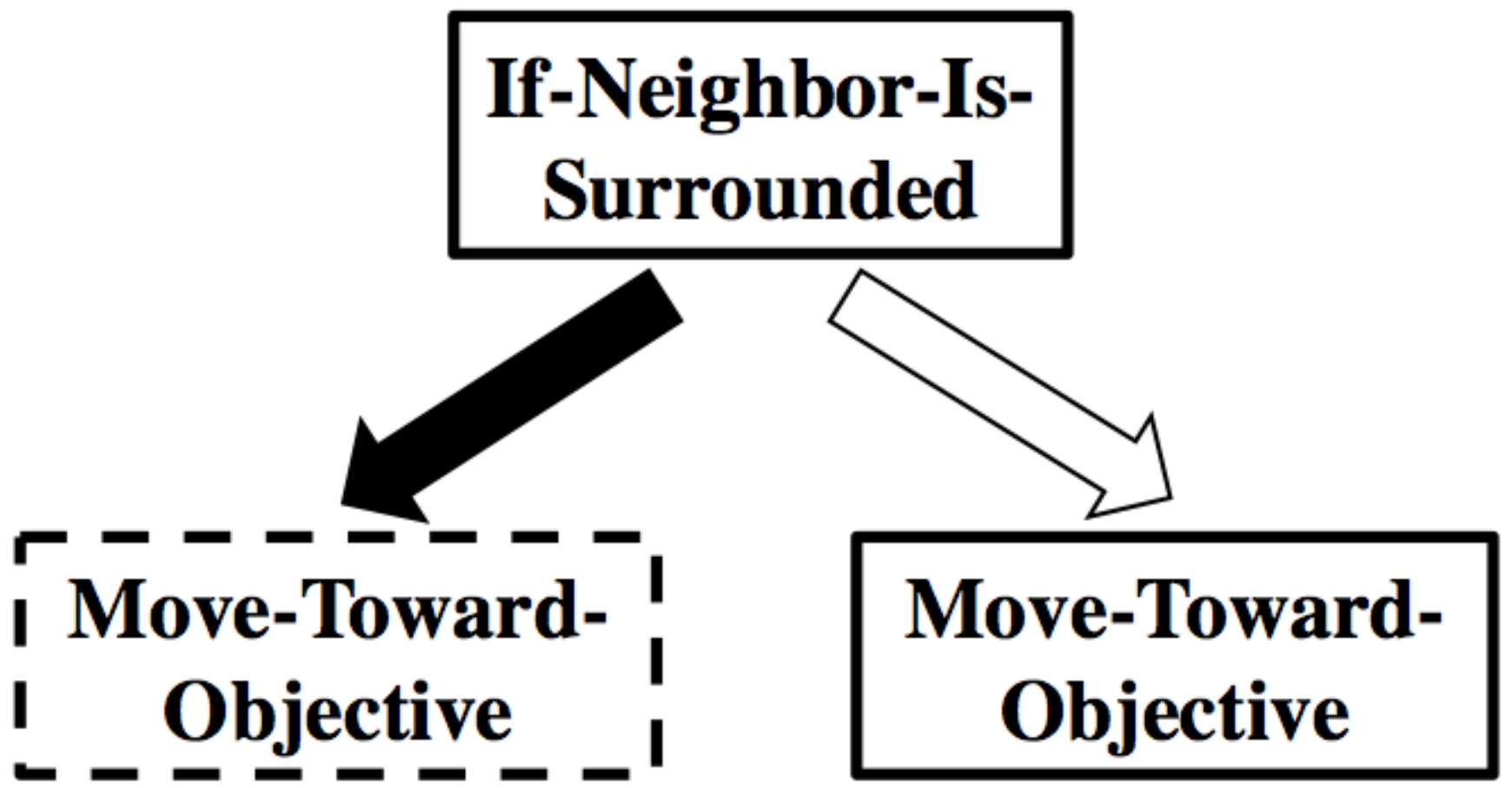}
\caption{}
\end{subfigure}
\begin{subfigure}[l]{0.5\textwidth}
\includegraphics[height=1.1in]{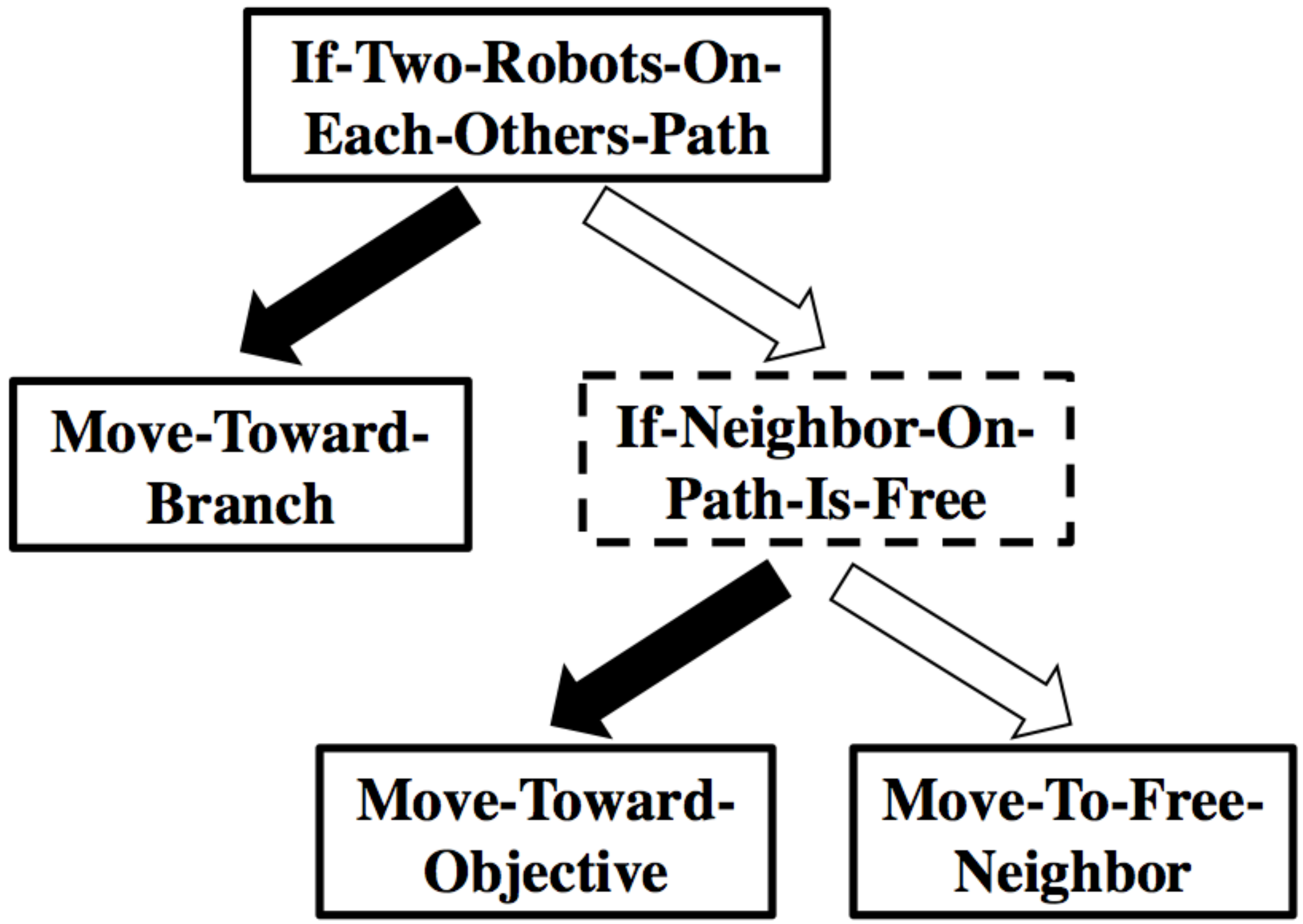}
\caption{}
\end{subfigure}
\caption{In (a) and (b), two example programs from a GP population are shown. In (c) and (d), crossover has been applied to the programs in (a) and (b) with crossover points shown via dashed boxes. (d) can also be used to demonstrate mutation, where a mutation point (dashed box) was replaced with a new randomly generated subtree.} \label{fig2}
\end{figure}

\section{Computational Complexity}
The genetic program returns a single path planning program $\psi*$ that is used by each robot within the environment. At each time step, the robots in the environment sequentially call $\psi*$ to determine their next action. As previously discussed, $\psi*$ has the form of a decision tree. An input parameter to the GP construction algorithm is $d_{max}$, dictating the maximum depth of $\psi*$. Thus, when a robot calls $\psi*$ and does a traversal of the decision tree, it traverses a single branch that consists of at most $d_{max}$ nodes. This means that at most $d_{max} - 1$ functions and 1 terminal are executed by each robot each time step.

To determine the computational complexity of having a single robot call $\psi*$ in a given time step, the most expensive function and terminal must be established because in the worst case, the most expensive function will be executed $d_{max} - 1$ times and the most expensive terminal will be executed once. The most expensive function is \texttt{If-Two-Robots-On-Each-Others-Path} and the most expensive terminal is \texttt{Move-Toward-Branch}.   

\texttt{If-Two-Robots-On-Each-Others-Path} has computational complexity $O(b^{\rho} + |N|^{2}log(|N|))$, where $b$ is the maximum branching factor in the tree $T(N,E)$ and all robots within $\rho$ nodes of a robot $r_i$ in $T(N,E)$ are considered to be in $r_i$'s direct communication network: $c(r_i)$. The $b^{\rho}$ term comes from having to construct the current robot's direct communication network $c(r_i)$. If two robots are on each other's paths, then the nearest branch node is found. This requires Dijkstra's algorithm to be called between $r_i$'s current position $A(r_i, t)$ and each branch node. The computational complexity of Dijkstra's algorithm is $O(E + Nlog(N))$ where $E$ and $N$ are the number of edges and nodes in the graph, respectively. Since the algorithm is being executed on a tree, $E = N - 1$. Thus, the computational complexity of Dijkstra's algorithm can be rewritten as $O(N + Nlog(N))$. The number of branch nodes is upper bounded by the number of nodes $|N|$. Therefore, finding the closest branch node to $r_i$ has complexity $O(|N|^{2}log(|N|))$.

\texttt{Move-Toward-Branch} also has computational complexity $O(|N|^{2}log(|N|))$, deriving from the requirement of finding the closest branch node to the current robot $r_i$.

In $O(b^{\rho} + |N|^{2}log(|N|))$, the maximum branching factor $b$ is typically much smaller than $|N|$ and in our experiments, $\rho$ was set to 2. Therefore, $b^{\rho} << |N|^{2}log(|N|)$. This leads us to conclude that the computational complexity of \texttt{If-Two-Robots-On-Each-Others-Path} and \texttt{Move-Toward-Branch} are each\\
$O(|N|^{2} log(|N|))$. The computational complexity of executing $\psi*$ once is, therefore, $O((d_{max} - 1)*(|N|^{2}log(|N|)) + |N|^{2}log(|N|))$ or $O(d_{max}|N|^{2}log(|N|))$.
 
\section{Developing the Function and Terminal Sets}
Developing the GP's function and terminal sets $F$ and $L$ for solving MRPP problems involved two phases. In phase one, functions and terminals enabling robots to swap positions were identified, a key requirement for removing deadlock in tunnel environments. The three unique MRPP scenarios requiring a swap were enumerated in \cite{references:PSW_thesis}. These scenarios have been adapted and described in terms of robots $r_A$ and $r_B$ here:
\begin{enumerate}
	\item $r_B$ is on the path from $r_A$ to $g(r_A)$ and $r_A$ is on the path from $r_B$ to $g(r_B)$
	\item Both $r_B$ and $g(r_B)$ are on the path from $r_A$ to $g(r_A)$
	\item $r_B$ is on the path from $r_A$ to $g(r_A)$ and $r_B$'s neighbors are all occupied by other robots
\end{enumerate}
	
These three scenarios were encoded in $X$ to train the GP. The sets $F$ and $L$ were fine-tuned until the GP was able to find solution programs.

In phase two, function set $F$ and terminal set $L$ were further fine-tuned until the GP was capable of consecutively solving 100 \textit{randomly} generated MRPP problems, each including a randomly constructed minimum spanning tree, occupied by a random number of robots having unique start and goal locations. To note, the number of robots was always fewer than the number of tree leaves (e.g. Fig. 1c). The final function set and terminal set can be found in Table 1 and Table 2, respectively.

\section{Experiments}
Three sets of experiments were conducted to evaluate GP's ability to generate programs for solving MRPP problems. The first set of experiments tested GP's ability to evolve programs for solving individual MRPP problems in the fewest possible time steps. The second set evaluated GP's capacity to evolve general programs for solving problems they were not trained on. The third set tested GP's ability to solve MRPP problems where PSW is not guaranteed to work.

In each experiment, a master computer first ran the GP algorithm offline on a fitness set of MRPP training examples. The GP algorithm was implemented using the EpochX Java library and was run on a Microsoft Surface Book with an Intel Core i7-6600U 71 CPU. The GP used a genetic crossover rate of $p_c = 0.8$, an asexual reproduction rate of $p_a = 0.1$, and a mutation rate of $p_m = 0.1$. Additionally, the algorithm ran for $i_r = 5$ runs of $i_g = 400$ generations. The population size was set to $|P| = 2000$. The program tree depths were bounded between 1 and 50. After the last run, the GP algorithm returned the learned program $\psi*$. Then, for each test problem, $\psi*$ was shared with each of the robots and executed online, in real-time, (as illustrated in Fig. \ref{figGPflow}). The robots sequentially executed $\psi*$ every time step to determine their next action until the problem was solved or the maximum number of allowable time steps had been reached.

In the plots below, each data point is the average y-value across examples sharing an x-value. Error bars represent the standard deviation of the y-value.

\subsection{Generating Training and Test Examples}
Training and test examples were recursively created with the same random minimum spanning tree (MST) generator shown in Algorithm 2. The generator was provided seed values between 4 and 10 to produce training examples and seed values between 4 and 9 to produce test examples. The maximum branching factor $b$ was set to 4. Once the MST was generated, it was populated with robots, each having a unique, random starting node and destination node. In the experiments described in sections 7.2 and 7.3, the number of robots was set to $\#leaves - 1$ because this is the maximum number of robots for which PSW is guaranteed to solve an environment. In section 7.4, the number of robots was set to $0.25$, $0.5$, $1.0$, and $1.5$ times the number of leaf nodes.

\begin{algorithm}
\begin{algorithmic}[1] 
  \caption{MSTGenerator(root, seed, b)}
    \STATE $n \leftarrow random(0, b)$\\
    \FOR{$i \leftarrow 1$ \TO $n$}
        \STATE $newNode \leftarrow createNode()$\\
        \STATE $MSTGenerator(newNode, seed - 1, b)$\\
        \STATE $root.addNeighbor(newNode)$\\
        \STATE $newNode.addNeighbor(root)$
    \ENDFOR
    \STATE \RETURN $root$
\end{algorithmic}
\end{algorithm}

\subsection{Optimizing for Time in Individual MRPPs}
GP programs were evolved to solve individual randomly generated MRPP examples in the fewest possible time steps. This process was conducted 1000 times. PSW was run against the same 1000 problems for comparison.

\begin{table}
\caption{GP vs. PSW for individual MRPP problems.}\label{tab1}
\begin{tabular}{
  @{\kern-.5\arrayrulewidth}
  |p{\dimexpr3.3cm-2\tabcolsep-.5\arrayrulewidth}
  |p{\dimexpr4.0cm-2\tabcolsep-.5\arrayrulewidth}
  |p{\dimexpr4.3cm-2\tabcolsep-.5\arrayrulewidth}
  |@{\kern-.5\arrayrulewidth}
}
\hline
{\bfseries GP Better} & {\bfseries GP = PSW} & {\bfseries PSW Better} \\
\hline
73.3$\%$ & 12.1$\%$ & 14.6$\%$ \\
\hline
\end{tabular}
\end{table}

The results presented in Table 3 show that GP outperformed or tied PSW in $85.4\%$ of the trials. Furthermore, PSW required 18633 time steps to solve the 1000 examples while GP only needed 8552 time steps to solve the 1000 examples, a $54.1\%$ improvement. Fig. 4 suggests that GP increasingly outperforms PSW as the number of robots navigating within an environment increases. Fig. 5 reveals that GP increasingly outperforms PSW as the $\#$ swaps/$\#$ robots ratio increases. In solving individual MRPP problems, then, GP significantly outperforms PSW and GP exhibits a greater capacity to scale as the number of robots navigating within an environment increases.

\begin{figure}
\includegraphics[width=\textwidth]{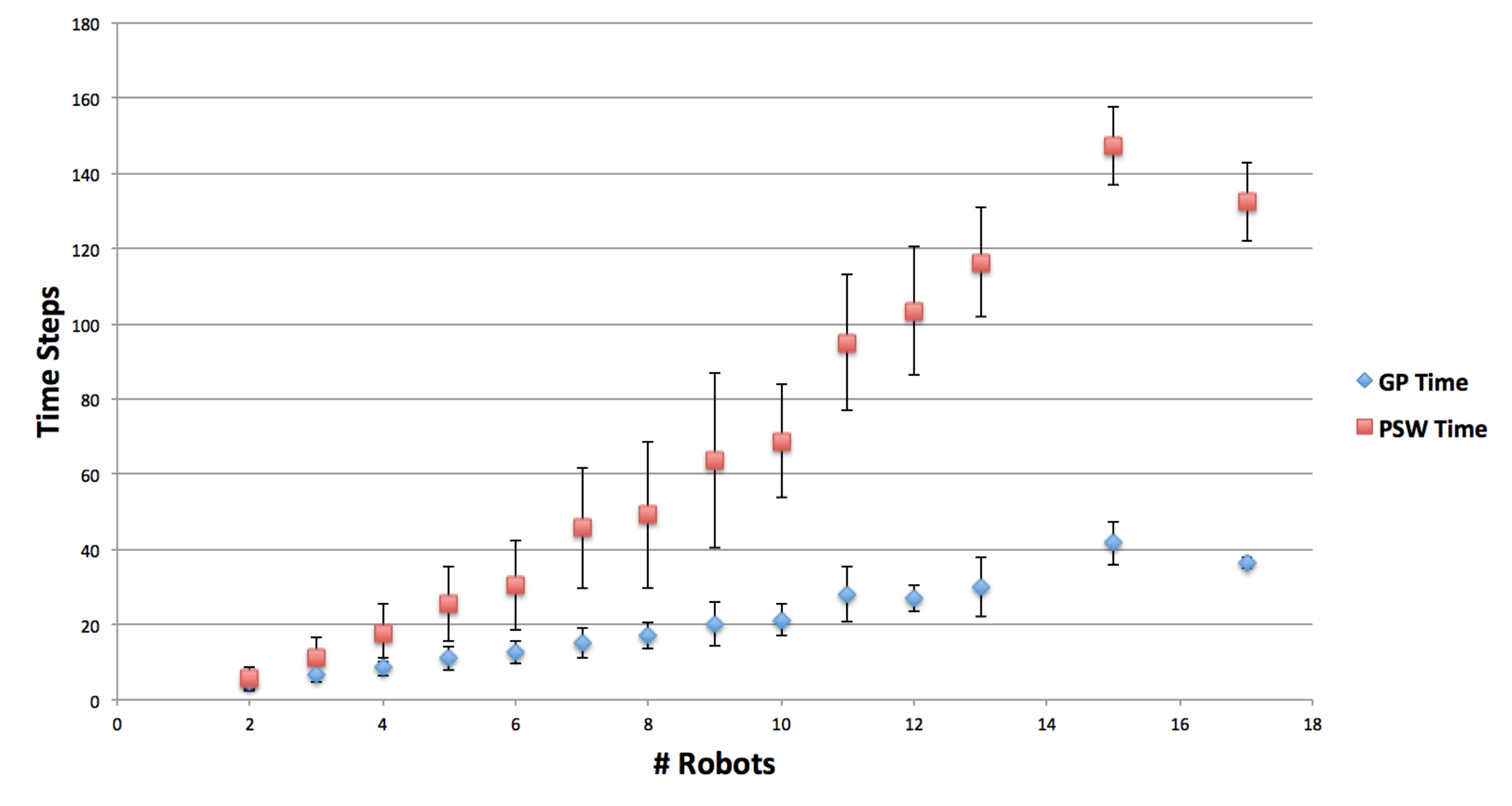}
\caption{GP vs PSW when trained/tested on individual MRPPs.} \label{fig9}
\end{figure}

\begin{figure}
\includegraphics[width=\textwidth]{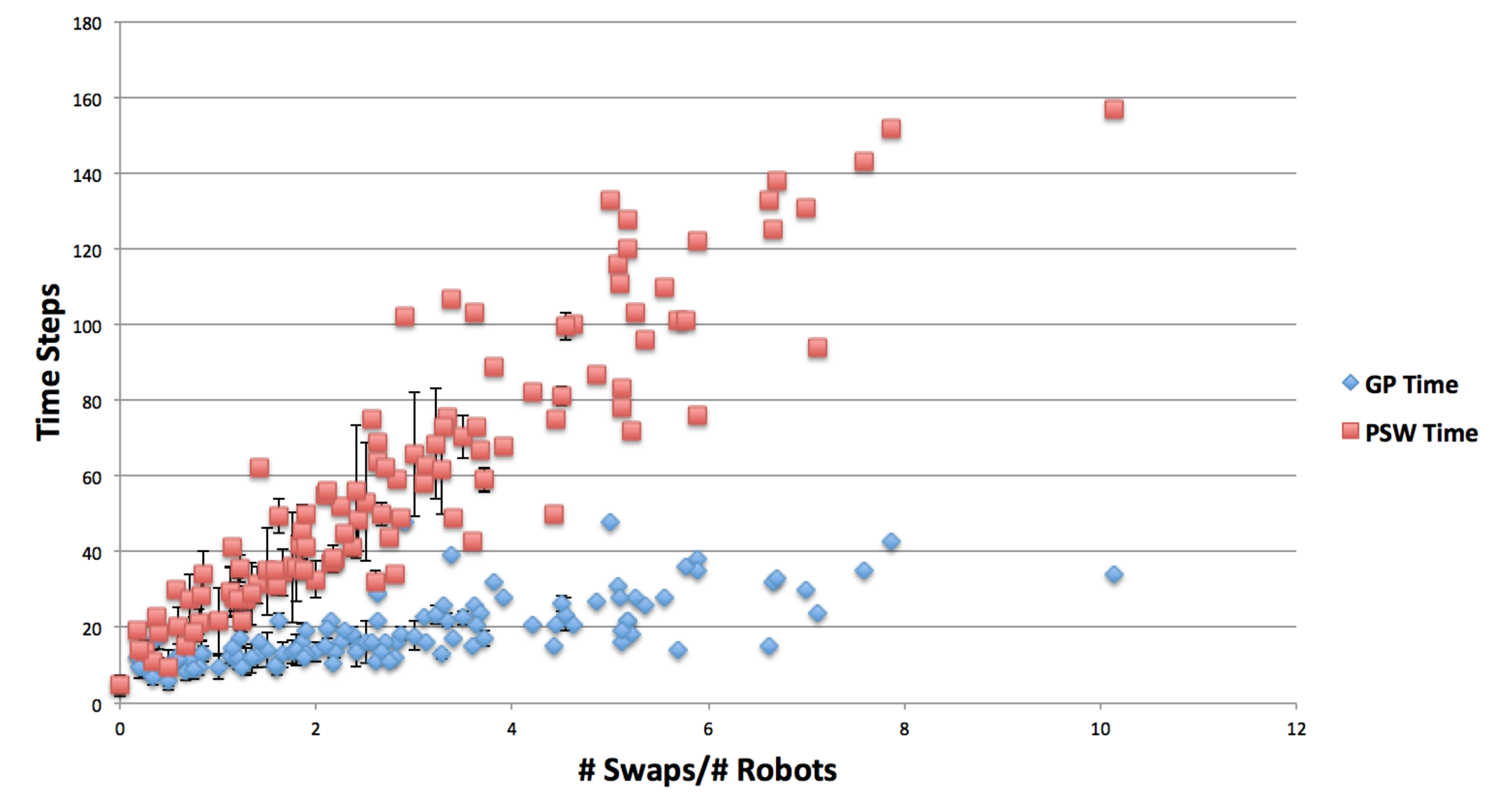}
\caption{GP vs PSW when trained/tested on individual MRPPs.} \label{fig10}
\end{figure}

\subsection{Generalizing  GP to Solve New MRPPs}
GP's capacity to evolve general MRPP programs for solving new problems, on which they were not trained, is examined next. Trials were conducted where the GP evolved based on a set of $|X|$ = 5, 10, or 20 randomly generated training problems. Next, the learned program was run against a test set of 100 randomly generated MRPP examples. This process was repeated ten times for each $|X|$ value (yielding 1000 examples for each $|X|$) and the results are presented in Table 4. 

These experiments reveal that GP's ability to produce general programs improves as $|X|$ increases. Despite the improvement, GP was not able to match PSW's completeness guarantee. Furthermore, in the problems solved by GP, increasing $|X|$ did not help GP outperform PSW as the percentage of examples in which GP required fewer time steps remained stable between $30.37\%$ and $36.69\%$. Fig. 6 reveals that PSW consistently outperforms GP programs over all problem sizes. However, Fig. 7 suggests that general GP programs outperform PSW when robots must, on average, perform more swaps.

\begin{figure}
\includegraphics[width=\textwidth]{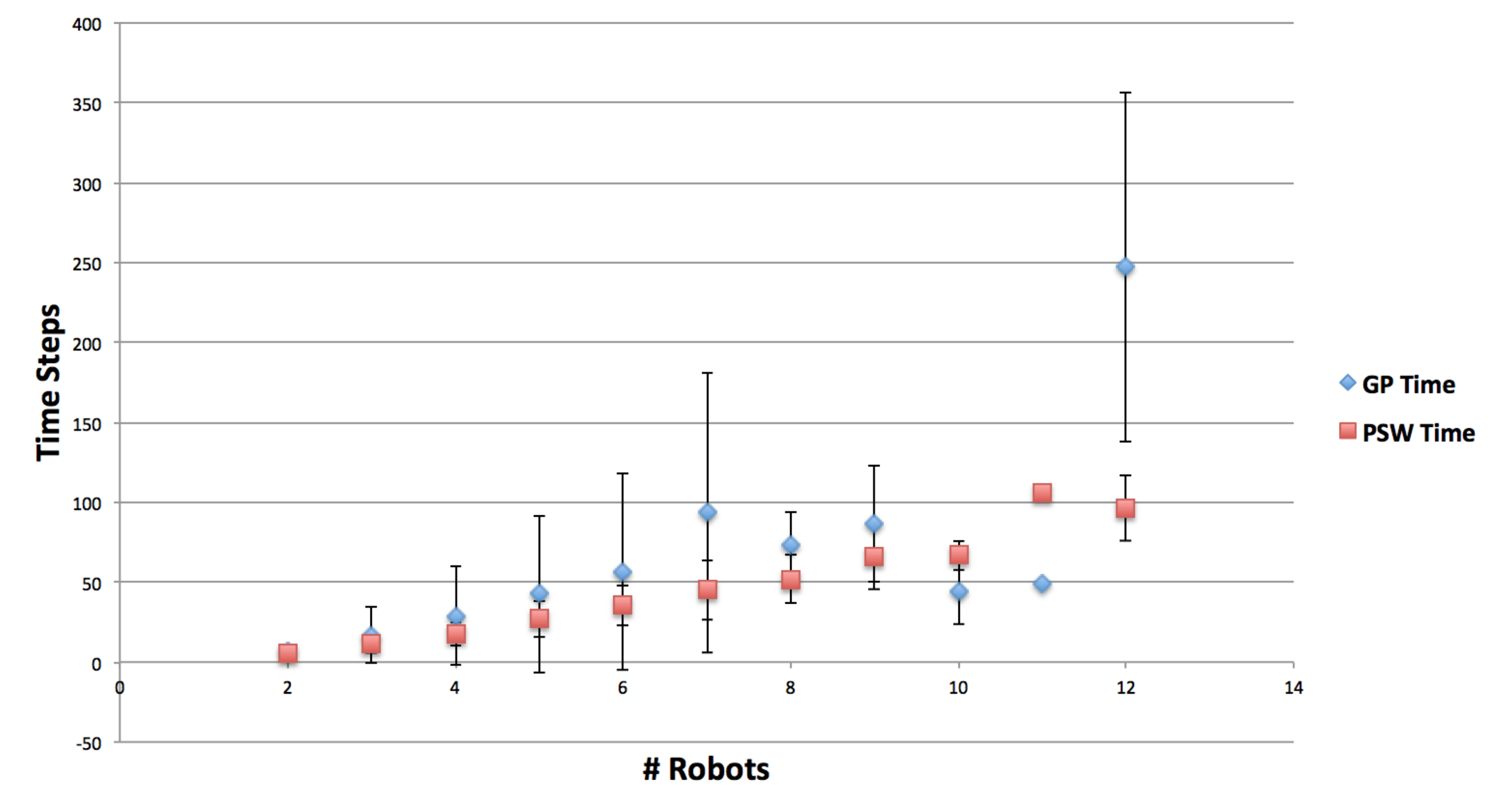}
\caption{GP vs PSW when trained first and then tested on NEW MRPPs.} \label{fig11}
\end{figure}

\begin{figure}
\includegraphics[width=\textwidth]{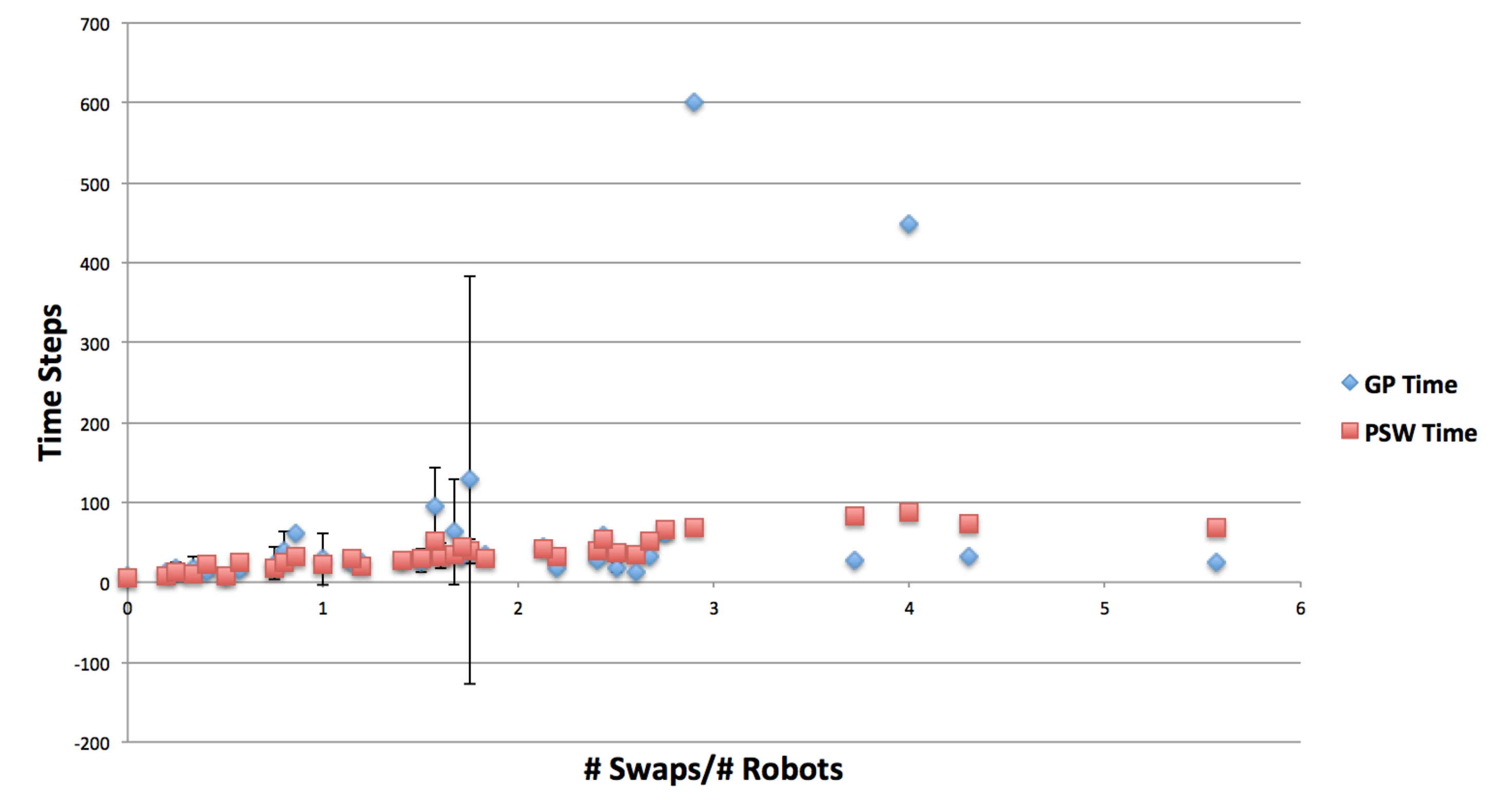}
\caption{GP vs PSW when trained first and then tested on NEW MRPPs.} \label{fig12}
\end{figure}

\begin{table}
\caption{General GP vs. PSW.}\label{tab1}
\begin{tabular}{
  @{\kern-.5\arrayrulewidth}
  |p{\dimexpr1.0cm-2\tabcolsep-.5\arrayrulewidth}
  |p{\dimexpr2.5cm-2\tabcolsep-.5\arrayrulewidth}
  |p{\dimexpr2.5cm-2\tabcolsep-.5\arrayrulewidth}
  |p{\dimexpr2.5cm-2\tabcolsep-.5\arrayrulewidth}
  |p{\dimexpr2.5cm-2\tabcolsep-.5\arrayrulewidth}
  |@{\kern-.5\arrayrulewidth}
}
\hline
{\bfseries $|X|$} & {\bfseries $\#$ GP Solved} & {\bfseries $\%$ GP Better} & {\bfseries $\%$ GP = PSW} & {\bfseries $\%$ PSW Better} \\
\hline
5 & 695 & 36.69 & 11.65 & 51.65 \\
\hline
10 & 866 & 30.37 & 15.24 & 54.39 \\
\hline
20 & 872 & 32.80 & 11.12 & 56.08 \\
\hline
\end{tabular}
\end{table}

\begin{table}
\caption{GP vs. PSW Where PSW Is Not Complete.}\label{tab1}
\begin{tabular}{
  @{\kern-.5\arrayrulewidth}
  |p{\dimexpr3.3cm-2\tabcolsep-.5\arrayrulewidth}
  |p{\dimexpr4.0cm-2\tabcolsep-.5\arrayrulewidth}
  |p{\dimexpr4.3cm-2\tabcolsep-.5\arrayrulewidth}
  |@{\kern-.5\arrayrulewidth}
}
\hline
{\bfseries Leaf Multiplier} & {\bfseries $\%$ GP Solved} & {\bfseries $\%$ PSW Solved} \\
\hline
0.25 & 100 & 100 \\
\hline
0.5 & 100 & 100 \\
\hline
1.0 & 82 & 68.5 \\
\hline
1.5 & 58 & 34 \\
\hline
\end{tabular}
\end{table}

\subsection{GP vs. PSW Where PSW Is Not Complete}
GP's ability to evolve programs solving individual MRPP problems violating PSW's completeness guarantee---examples where $|R| \geq \#leaves$---is examined next. Here, 200 trials were conducted in which the number of robots was fixed at $|R| = \floor{l_m*\#leaves}$, where $l_m$ was set to 0.25, 0.5, 1.0, and 1.5, respectively. In each trial, a GP program was evolved to solve an individual randomly generated MRPP example and PSW was run against the same problem for comparison. If $|R| > |N| - 2$, $|R|$ was set to $|N| - 2$ because at least two nodes must be free to enable a swap. In the trials in which PSW's completeness guarantee was in effect, both PSW and GP were able to find solutions $100\%$ of the time (see Table 5). When PSW's completeness guarantee was violated ($|R| = \# leaves$ and $|R| = \floor{1.5*\# leaves}$), GP consistently solved more examples than PSW, e.g. solving $24\%$ more examples when $|R| = \floor{1.5*\# leaves}$.

\section{Conclusions and Future Work}
This paper presents a decentralized, local GP approach for solving single-lane MRPP problems. This GP approach facilitates the generation of MRPP programs that can optimize for various attributes and scenarios. GP effectively optimizes the number of time steps needed to solve individual MRPP problems, showing significant improvement over complete algorithms like PSW. GP also consistently outperforms PSW in solving problems that do not meet PSW's completeness conditions. Furthermore, GP exhibits a greater capacity than PSW to scale as the number of robots navigating within an MRPP environment increases. This research illustrates the benefits of using GP to generate solutions for individual MRPP problems, including instances in which the number of robots exceeds the number of leaves in the tree-modeled workspace. However, attempts to use GP to produce general MRPP programs could not match PSW's completeness despite the suggestion that general GP programs do outperform PSW where robots must perform more swaps. Future work should focus on developing functions and terminals enabling GP to produce more general MRPP programs and programs that better optimize for time. Additional experiments should reduce the constraints on the environment by training and testing on general graphs rather than MSTs. Furthermore, future research should attempt to modify this GP framework so that the learned program can be executed asynchronously within the environment.  Finally, the programs produced by GP should be implemented on physical robots to better understand the nature of these programs and to observe where there is room for improvement.

%
%

%
%
%
%

\bibliographystyle{splncs04}
\bibliography{main}

\end{document}